\documentclass[10pt,twocolumn,letterpaper]{article}

\usepackage{cvpr}              

\usepackage{graphicx}
\usepackage{amsmath}
\usepackage{amssymb}
\usepackage{booktabs}
\usepackage{subfiles}
\usepackage{caption}
\usepackage{subcaption}
\usepackage{multirow}
\usepackage[pagebackref,breaklinks,colorlinks]{hyperref}

\usepackage[capitalize]{cleveref}
\crefname{section}{Sec.}{Secs.}
\Crefname{section}{Section}{Sections}
\Crefname{table}{Table}{Tables}
\crefname{table}{Tab.}{Tabs.}

\def\cvprPaperID{*****} 
\def\confName{CVPR}
\def\confYear{2022}

\begin{document}

\title{UniFusion: Unified Multi-view Fusion Transformer for Spatial-Temporal\\ Representation in Bird's-Eye-View}

\author{Zequn Qin$^{1}$, Jingyu Chen$^2$, Chao Chen$^2$, Xiaozhi Chen$^{2}$\thanks{Corresponding authors: Xiaozhi Chen and Xi Li.}, Xi Li$^{1,3,4}$\footnotemark[1]
\\
$^1$College of Computer Science, Zhejiang University; $^2$DJI
\\
$^3$Shanghai Institute for Advanced Study of Zhejiang University; $^4$Shanghai AI Lab
\\
{\tt \small zequnqin@gmail.com, jeffery.chen@dji.com, huaijin.chen@dji.com}\\
{\tt \small cxz.thu@gmail.com, xilizju@zju.edu.cn}
}

\maketitle

\begin{abstract}
Bird's eye view (BEV) representation is a new perception formulation for autonomous driving, which is based on spatial fusion. Further, temporal fusion is also introduced in BEV representation and gains great success. In this work, we propose a new method that unifies both spatial and temporal fusion and merges them into a unified mathematical formulation. The unified fusion could not only provide a new perspective on BEV fusion but also brings new capabilities. With the proposed unified spatial-temporal fusion, our method could support long-range fusion, which is hard to achieve in conventional BEV methods. Moreover, the BEV fusion in our work is temporal-adaptive and the weights of temporal fusion are learnable. In contrast, conventional methods mainly use fixed and equal weights for temporal fusion. Besides, the proposed unified fusion could avoid information lost in conventional BEV fusion methods and make full use of features. Extensive experiments and ablation studies on the NuScenes dataset show the effectiveness of the proposed method and our method gains the state-of-the-art performance in the map segmentation task.
\end{abstract}
\vspace{-20pt}
\section{Introducion}
\label{sec_intro}
\begin{figure}
    \centering
    \begin{subfigure}[b]{0.57\linewidth}
        \centering
        \includegraphics[width=\textwidth]{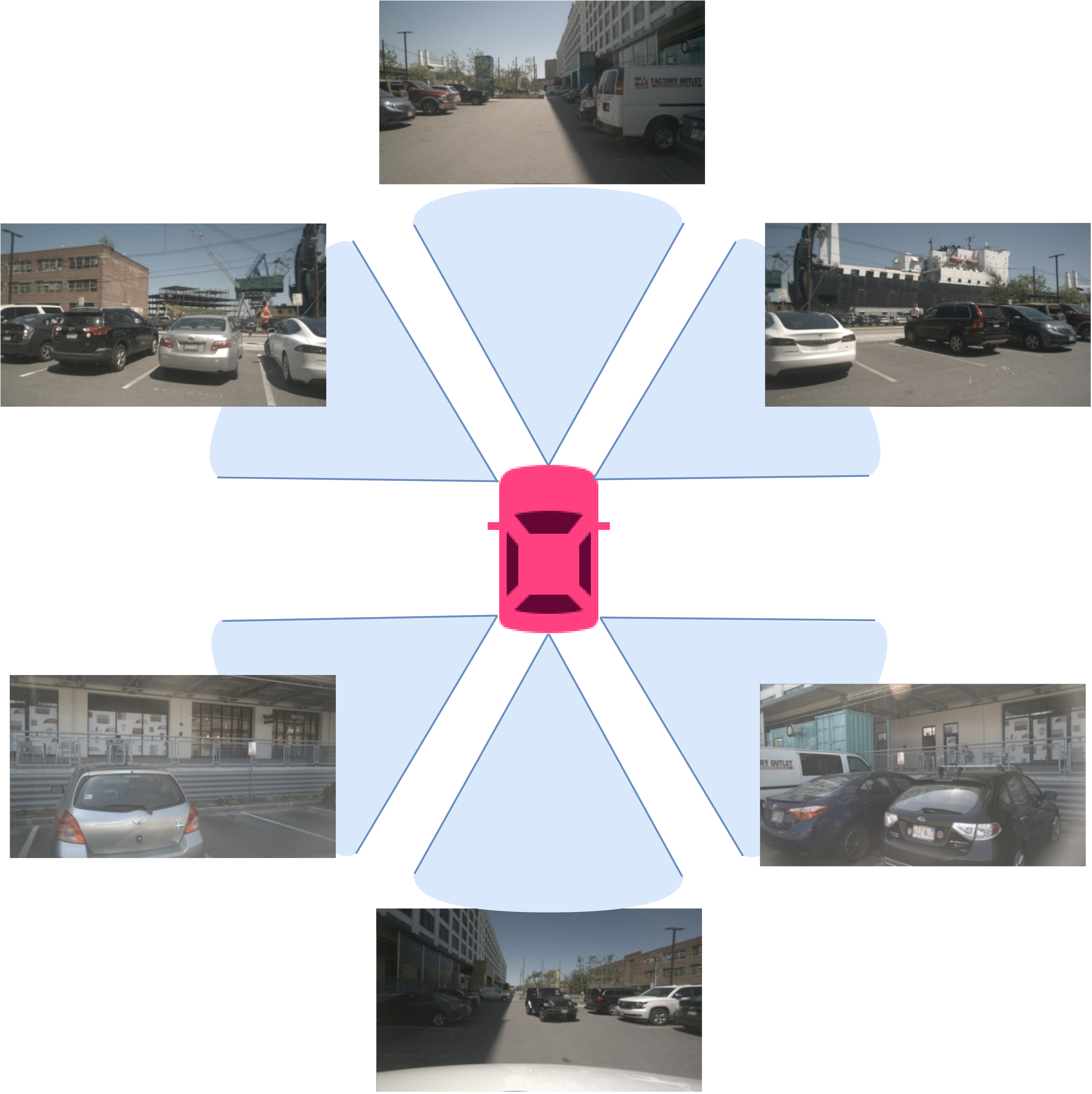}
        \caption{Inputs with surrounding images.}
        \label{fig_intro_1}
    \end{subfigure}
    \hfill
    \begin{subfigure}[b]{0.27\linewidth}
        \centering
        \includegraphics[width=\textwidth]{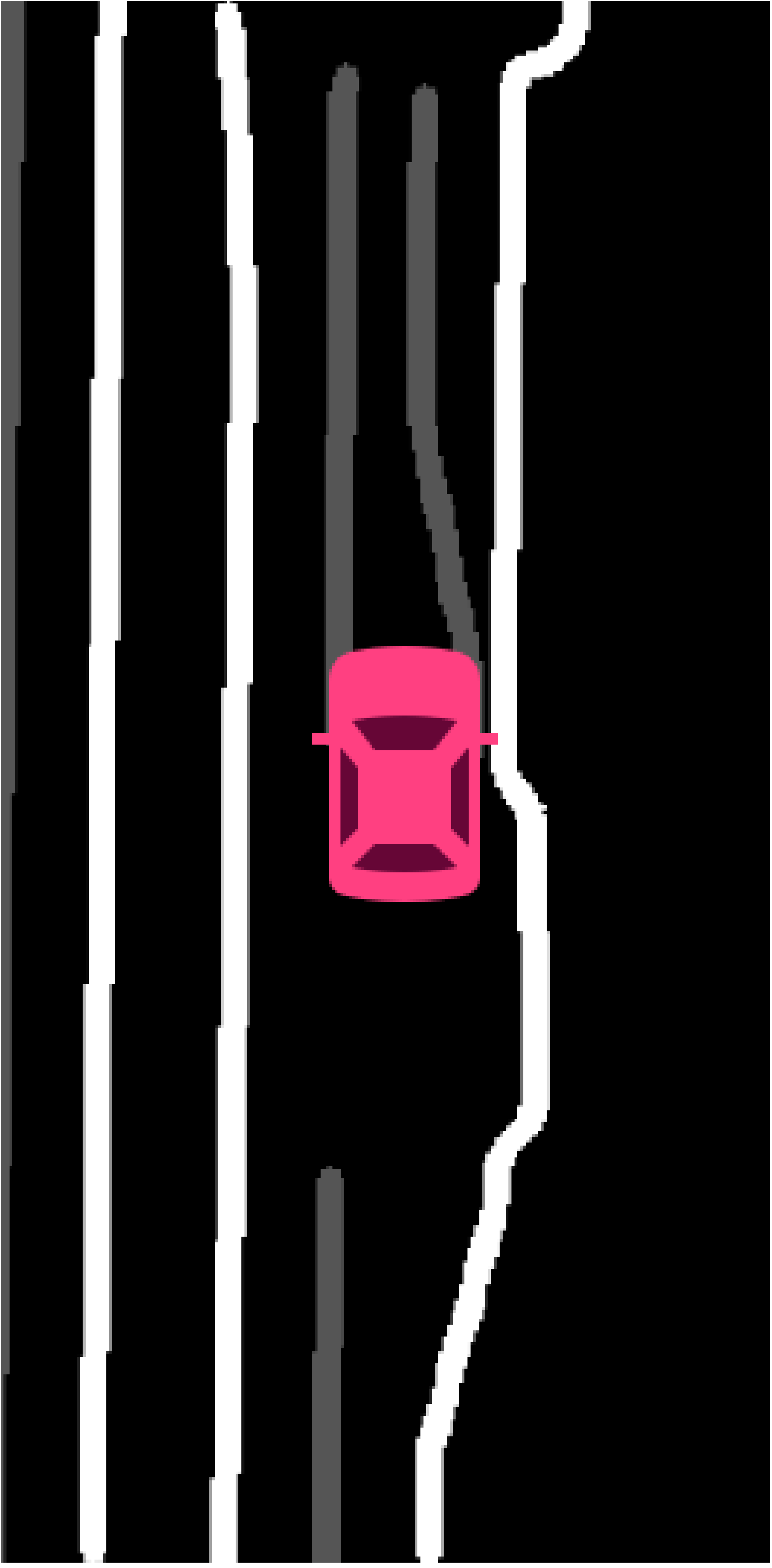}
        \caption{Map.}
        \label{fig_intro_2}
    \end{subfigure}
    \caption{Illustration of the map segmentation task in BEV.}
    \label{fig_intro}
    \vspace{-10pt}
\end{figure}

Recently, bird's-eye-view (BEV) representation~\cite{pan2020cross,philion2020lift,li2021hdmapnet} becomes an emerging perception formulation in the autonomous driving field. The main idea of BEV representation is to map the multi-camera features into the ego BEV space, \ie, spatial fusion, as shown in \cref{fig_intro}. This kind of spatial fusion composes an integrated BEV space, and duplicate results from different cameras are uniquely represented in the BEV space, which greatly reduces the difficulty in fusing multi-camera features. Moreover, the BEV spatial fusion naturally shares the same 3D space as other modalities like LiDAR and radar, making multi-modality fusion simple.

The integrated BEV representation based on spatial fusion provides the basis of temporal fusion. Temporal fusion is a cornerstone in BEV representation, which can be used in many aspects like 1) representing temporarily occluded objects; 2) accumulating observation in a long-range, which can be used for generating map; 3) stabilizing the perception results for standstill vehicles. There have been many methods~\cite{li2021hdmapnet,huang2022bevdet4d,li2022bevformer} showing the importance and effectiveness of temporal fusion. 

\begin{figure*}
    \centering
    \includegraphics[width=\textwidth]{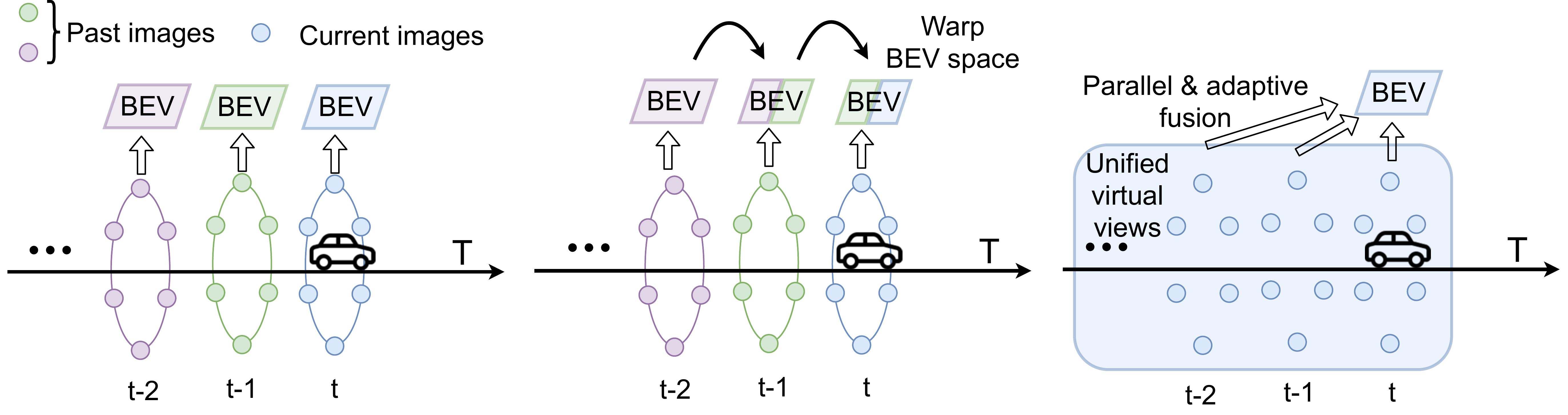}
    \caption{Different methods in BEV temporal fusion. From left to right, they are methods with no temporal fusion, warp-based temporal fusion, and our unified multi-view fusion. For the method with no temporal fusion, the BEV space is only predicted with surrounding images at the current time step. The warp-based temporal fusion would warp the BEV space from the previous time step and is a serial fusion method. In this work, we propose unified multi-view fusion, which is a parallel method and could support long-range fusion.}
    \label{fig_intro_diff}
    \vspace{-10pt}
\end{figure*}

Despite the success of current progress, present methods usually use warp-based temporal fusion, \ie, warping past BEV features to the current time according to the positions of BEV spaces at different time steps. Although this kind of design can well align temporal information, there are still some open problems. First, the warping is usually serial; that is to say, it is conducted only between adjacent time steps. In this way, it is hard to model long-range temporal fusion. Long-range history information can only implicitly make an impact and would be forgotten and dispelled rapidly. Besides, excessive long temporal fusion would even harm the performance in the warp-based temporal fusion. Second, warping would cause information loss during temporal fusion, as shown in \cref{fig_intro_vv_1,fig_intro_vv_2}. Third, since the warping is serial, the weights for all time steps are equal, and it is hard to adaptively fuse temporal information.

To solve the above problems, we propose a new perspective that combines both spatial and temporal fusion into a unified multi-view fusion, termed UniFusion. Specifically, spatial fusion is regarded as a multi-view fusion from multi-camera features. For the temporal fusion, since the temporal features are from the past and absent in the current time, we create ``virtual views'' for the temporal features as if they are present in the current time. 
The idea of ``virtual views'' is to treat past camera views as the current views and assign them virtual locations relative to the current BEV space based on the camera motion.
In this way, the whole spatial-temporal representation in BEV can be simply treated as a unified multi-view fusion, which contains both current (spatial fusion) and past (temporal fusion) virtual views, as shown in \cref{fig_intro_diff}.

With the proposed unified fusion, both spatial and temporal fusions are conducted in parallel. We can directly access all useful features through space and time at once, which enables the long-range fusion. Another benefit is that we can realize adaptive temporal fusion since we can directly access all temporal features. Meanwhile, the parallel property guarantees that no information is lost during fusion. Furthermore, the multi-view unified fusion can even support different sensors, camera rigs, and camera types at different time steps. This will bridge higher-level and heterogeneous fusion like vehicle-side and road-side perceptions. For example, we can fuse information from a car's camera and a surveillance camera on top of a traffic light, as long as they overlap in the BEV space.

The contributions of this work are as follows:
\begin{itemize}
    \vspace{-5pt}
    \item We propose a new parallel multi-view perspective for BEV representation, which unifies the spatial and temporal fusion. The proposed unified parallel multi-view fusion can address the problem of long-range fusion and information loss. And we can realize adaptive temporal fusion based on the unified fusion. The proposed unified method can also support arbitrary camera rigs and bridge higher-level and heterogeneous fusion.
    \vspace{-5pt}
    \item We analyze the widely used evaluation settings in the map segmentation task on NuScenes~\cite{caesar2020nuscenes} and propose a new setting for a more comprehensive comparison in \cref{sec_eval}.
    \vspace{-5pt}
    \item The proposed method achieves the state-of-the-art BEV map segmentation performance on the challenging benchmark NuScenes in all settings.
\end{itemize}
\section{Related Work}
\paragraph{Spatial fusion in BEV}
Spatial fusion is the basis of BEV representation, \ie, how to transform and fuse information and features from surrounding multi-camera inputs into an ego BEV space to represent the surrounding 3D world. The earliest and most straightforward method is the inverse perspective mapping (IPM)~\cite{matthies1992stereo, bertozzi1996real, aly2008real, deng2019restricted}, which assumes the ground surface is flat and at a fixed height. In this way, the spatial fusion in BEV can be conducted with a homography transformation. Note that IPM is usually utilized in the image space. However, IPM is hard to cope with the non-flat and unknown-height ground surface. Later, View Parsing Network (VPN)~\cite{pan2020cross} uses a fully connected layer to transform the image features into the BEV features and directly supervise the features in the BEV space in an end-to-end manner. Similarly, BEVSegFormer~\cite{peng2022bevsegformer} uses the deformable attention~\cite{zhu2020deformable} mechanism to achieve end-to-end mapping. These methods avoid the explicit mapping between image and BEV spaces, but this property also makes them hard to adopt the geometry prior. Based on VPN, HDMapNet~\cite{li2021hdmapnet} proposes to only map the image space to camera-ego BEV space in an end-to-end manner, while the multi-camera BEV spaces are fused with the camera poses. In this way, part of the geometry prior, \ie, the camera extrinsic information is utilized. 
To make full use of geometry prior in the spatial fusion of BEV space, Lift-splat-shoot~\cite{philion2020lift} proposes a latent estimation network to predict depth for each pixel in the image space. Then all the pixels with depth can be directly mapped into the BEV space. Another kind of method OFT~\cite{roddick2018orthographic} does not make predictions of depth. OFT directly copy-and-paste the features in the image space to all locations that trace along the ray from the camera in the BEV space. Different from the spatial fusion perspective of geometric mapping, X-Align\cite{borse2023x} aligns the semantics of camera and BEV spaces.
\vspace{-10pt}
\paragraph{Temporal fusion in BEV}
With the basis of spatial fusion, temporal fusion could further boost the representation in BEV space. The mainstream methods of temporal fusion are the warp-based method~\cite{zhang2022beverse,huang2022bevdet4d,li2022bevformer}. The main idea of the warp-based method is to warp and align BEV spaces at different time steps based on the ego motions of vehicles. The major differences reflect in the way of using wrapped BEV spaces. BEVFormer~\cite{li2022bevformer} uses deformable self-attention to fuse wrapped BEV spaces while BEVDet4D directly concatenates the wrapped BEV spaces. BEVFusion proposes~\cite{liu2022bevfusion} a unified multi-task and multi-sensor fusion method that can fuse camera and LIDAR. 
\section{Method}
In this section, we elaborate on the design of our method from two aspects. First, we show the derivation of the unified multi-view fusion. Then we demonstrate the network architecture with unified multi-view fusion.

\subsection{Unified Fusion with Virtual Views}

As discussed in the introduction, spatial fusion is the foundation of BEV representation, while temporal fusion reveals a new direction for better BEV representation.

\begin{figure}
    \centering
    \begin{subfigure}[b]{0.75\linewidth}
        \centering
        \includegraphics[width=\textwidth]{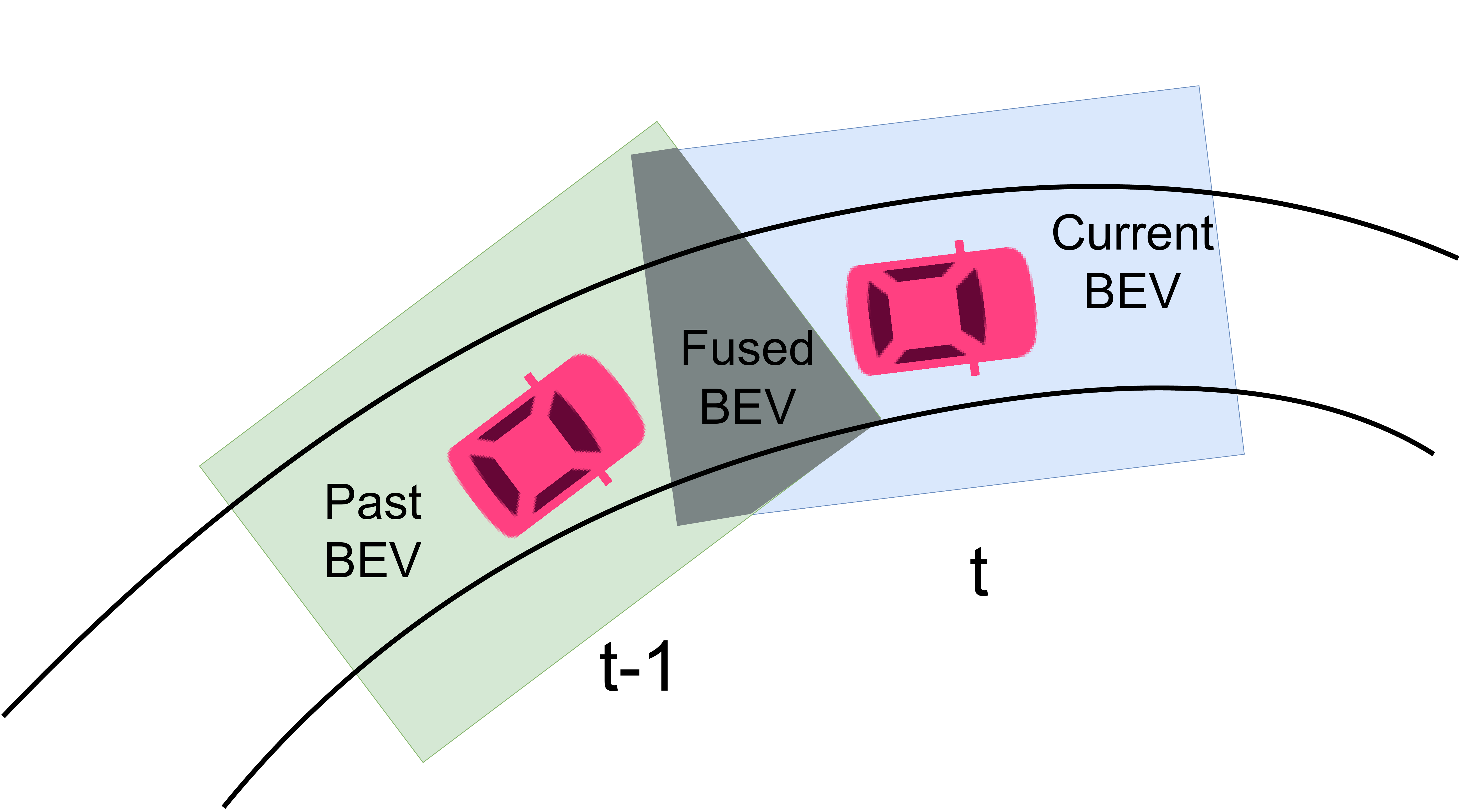}
        \caption{Warp-based BEV fusion. The fused area is marked in gray.}
        \label{fig_intro_vv_1}
    \end{subfigure}
    \hfill
    \begin{subfigure}[b]{0.75\linewidth}
        \centering
        \includegraphics[width=\textwidth]{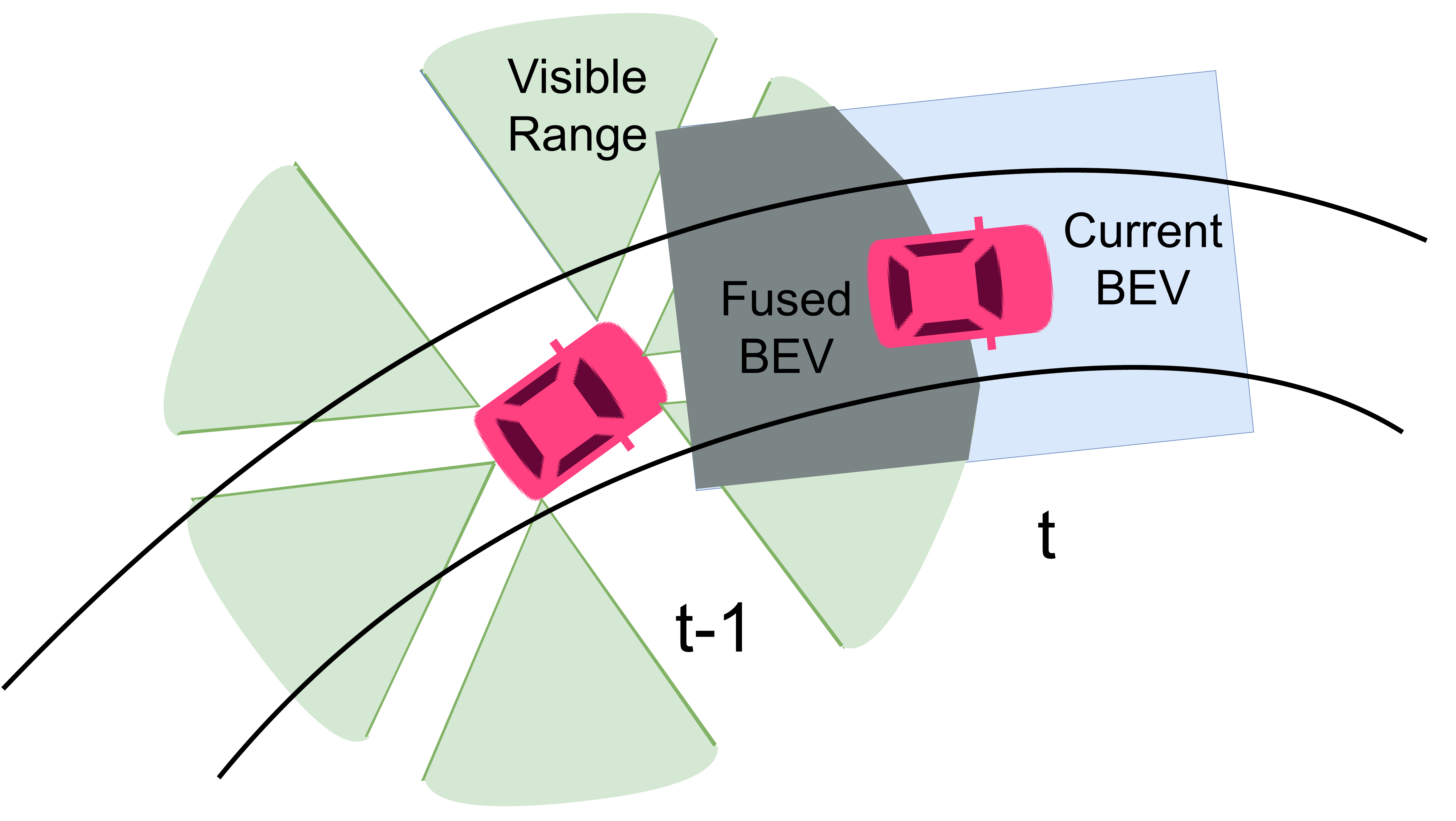}
        \caption{Actual BEV space that can be fused.}
        \label{fig_intro_vv_2}
    \end{subfigure}
    \begin{subfigure}[b]{0.75\linewidth}
        \centering
        \includegraphics[width=\textwidth]{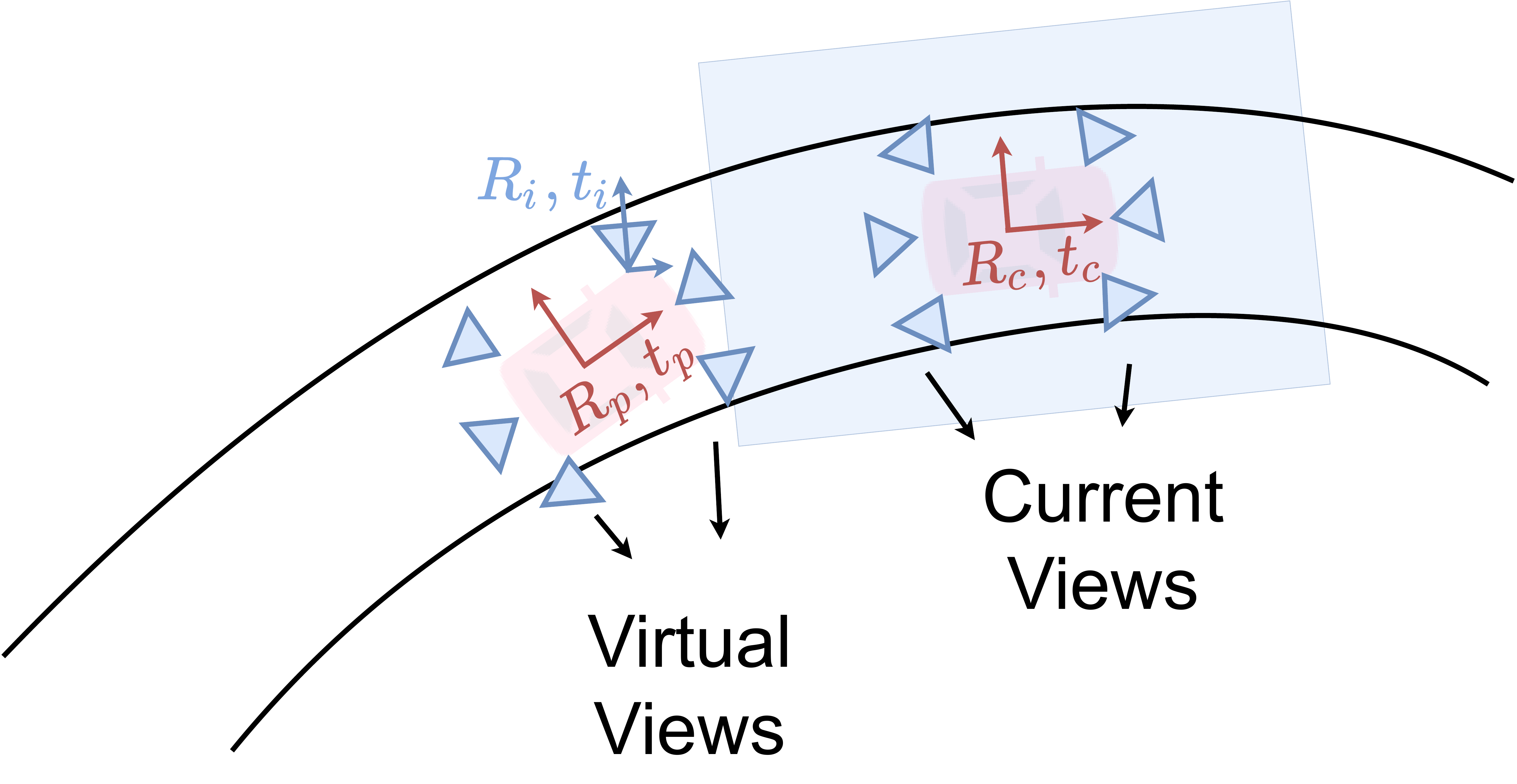}
        \caption{Illustration of virtual views.}
        \label{fig_intro_vv_3}
    \end{subfigure}
    \caption{Derivation of virtual views.}
    \label{fig_intro_vv}
    \vspace{-10pt}
\end{figure}

Conventional BEV temporal fusion is warp-based fusion, as shown in \cref{fig_intro_vv_1}. The warp-based fusion warps past BEV features and information based on the ego-motion of different time steps. Since all features are already organized in a pre-defined ego BEV space at a certain time step before warping, this process would lose information. 

The actual visible range of a camera is much bigger than the one of ego BEV space. For example, 100m is a very humble visible range for typical cameras, while most BEV ranges are defined as no more than 52m~\cite{li2022bevformer,philion2020lift}. In this way, it is possible to obtain better BEV temporal fusion than simply warping BEV spaces, as shown in \cref{fig_intro_vv_2}.

To achieve better temporal fusion, we propose a new concept, \ie, virtual view, as shown in \cref{fig_intro_vv_3}. Virtual views are defined as the views of sensors that do not present in the current time step, and these past views are rotated and translated according to the ego BEV space as if they are present in the current time step. Denote $R_c\in \mathbb{R}^{3 \times 3}, t_c\in \mathbb{R}^{3 \times 1}$ and $R_p \in \mathbb{R}^{3 \times 3}, t_p\in \mathbb{R}^{3 \times 1}$ as the rotations and translations matrices of current and past ego BEV spaces, respectively. Suppose $R_i\in \mathbb{R}^{3 \times 3}$, $t_i\in \mathbb{R}^{3 \times 1}$, and $K_i\in \mathbb{R}^{3 \times 3}$ are the rotation, translation and intrinsic matrices of a certain view $V_i$. The rotation and translation matrices of virtual views can be written as:

\begin{equation}
    \begin{aligned}
        R_i^v =& R_i^{-1} R_p^{-1} R_c \\
        t_i^v =& R_i^{-1} R_p^{-1} t_c - R_i^{-1} R_p^{-1} t_p - R_i^{-1} t_i,
    \end{aligned}
    \label{eq_uni_rot_t}
\end{equation}
in which $R_i^v\in \mathbb{R}^{3 \times 3}$ and $t_i^v\in \mathbb{R}^{3 \times 1}$ are the unified virtual rotation and translation matrices for any view $V_i$. It can be examined that \cref{eq_uni_rot_t} also holds for the current views. In this way, all views can be mapped and utilized in the same way, no matter they are past or current views. Suppose $P_{bev}\in \mathbb{R}^{N \times 3}$ represents the coordinates in the BEV space, $P_{img}\in \mathbb{R}^{N \times 3}$ is the homogeneous coordinates in the image space, and $N$ is the number of coordinates. The mapping between BEV space and all views can be written as:
\begin{equation}
    P_{img} = K_i (R_i^v P_{bev} + t_i^v).
    \label{eq_map}
\end{equation}
Then we can map the image features to the BEV features $F$.

\begin{figure*}[t]
    \centering
    \includegraphics[width=0.95\linewidth]{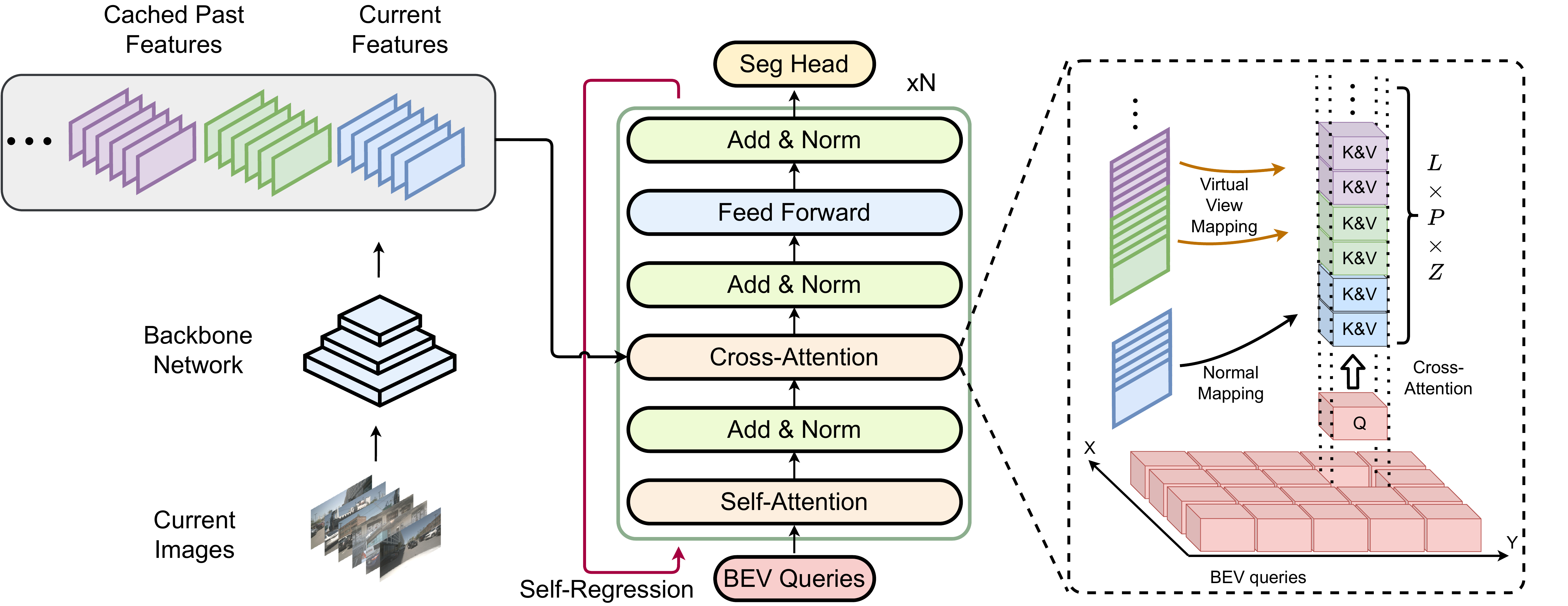}
    \caption{Network architecture.}
    \label{fig_main}
    \vspace{-10pt}
\end{figure*}

\subsection{Network Design with Unified Fusion}
\label{sec_net}
With the help of the unified multi-view fusion, we show the network architecture in this part. The network is composed of three parts, which are the backbone network, unified multi-view fusion Transformer, and segmentation head, as shown in \cref{fig_main}. 
\vspace{-10pt}
\paragraph{Backbone} We use three kinds of widely used backbones ResNet50~\cite{he2016deep}, Swin-Tiny~\cite{liu2021swin} and VoVNet~\cite{lee2019energy} to extract $L$ multi-scale features ($L = 4$) from multi-camera images. For the ResNet50 and VoVNet models, only features from stages 2, 3, and 4 are used. Following Deformable-DETR~\cite{zhu2020deformable}, an extra 3x3 convolution with a stride of 2 is used to generate the last feature. The backbone is shared between all views' images. It is worth mentioning that the features of past images can be maintained and reused in a feature queue without extra computational cost.
\vspace{-10pt}
\paragraph{Fusion Transformer}
We use a Transformer~\cite{NIPS2017_3f5ee243} encoder to fusion features from all views. There are four major parts in the Transformer encoder, which are the BEV queries, the self-attention module, the cross-attention model, and the self-regression mechanism.

In order to represent the BEV space, we use $X \times Y$ queries $\{Q_{x,y} \in \mathbb{R}^C | x \in \{1, \cdots, X\}, y \in \{1, \cdots, Y\}\}$ in a 2D grid to represent the whole BEV space, where $X$ and $Y$ are the spatial sizes of the BEV grid.

The second major part is the self-attention module. It is used to interact with all BEV queries and exchange information in the BEV space. Since the time complexity of the vanilla self-attention interaction is $O(X^2Y^2)$, we use deformable self-attention~\cite{zhu2020deformable} to reduce the computational cost.

The most important module of this work is the cross-attention used for unified multi-view spatial-temporal fusion. With the help of the unified multi-view fusion, all spatial-temporal features can be mapped to the same ego BEV space. The goal of the cross-attention module is to fuse and integrate the mapped spatial-temporal BEV space features $F$. 

Denote $(\hat{x},\hat{y},\hat{z})$ are the real-world coordinates in the 2D BEV grid $(x,y)$, and $\hat{z}$ is the real-world height for sampling. Suppose the number of sampling in height in each BEV grid is $Z$, then each BEV query $Q_{x,y}$ corresponds to $Z$ points, and the total coordinates in the BEV space is $P_{bev}\in \mathbb{R}^{XYZ \times 3}$. Then we can obtain the mapped BEV features $F$ according to \cref{eq_map} with $P_{bev}$. Suppose the number of time steps in temporal fusion is $P$, then the cross-attention (CA) module can be written as:
\begin{equation}
    \text{CA}(Q_{x,y}, F) = \sum_{p,l,z} \dfrac{e^{att_{x,y}^{p,l,z}}}{\sum_{p,l,z} e^{att_{x,y}^{p,l,z}}} F_{x,y}^{p,l,z},
    \label{eq_ca}
\end{equation}
where $F_{x,y}^{p,l,z}$ is the sampled value at the point of $(\hat{x},\hat{y},\hat{z})$ from the BEV features $F$ of $l$-th multi-scale level and $p$-th time step. $\sum_{p,l,z}$ is the summation over $P$ time steps, $L$ scales, and $Z$ heights. The attention value of $att_{x,y}^{p,l,z}$ is:
\begin{equation}
    att_{x,y}^{p,l,z} = \dfrac{Q_{x,y}K_{x,y}^{p,l,z}}{\sqrt{C}},
\end{equation}
in which $C$ is the dimension of each BEV query, and $K_{x,y}^{p,l,z}$ is the attention key composed of input $F_{x,y}^{p,l,z}$ and positional embedding.

\begin{table*}
    \centering
    \caption{Comparison of different map segmentation settings on NuScenes.}
    \label{tb_setting}
    \begin{tabular}{ccccccc}
    \toprule
    Setting     & Front/rear range & Left/right range & BEV grid size & Map element type & Line width & Split      \\ \midrule
    100m $\times$ 100m & 50m              & 50m              & 0.5m $\times$ 0.5m   & Line, polygon     & 1-pixel    & Vanilla    \\
    60m $\times$ 30m   & 30m              & 15m              & 0.15m $\times$ 0.15m & Line             & 5-pixel    & Vanilla    \\
    160m $\times$ 100m  & 100m/60m          & 50m              & 0.25m $\times$ 0.25m & Line             & 3-pixel    & City-based \\
    \bottomrule
    \end{tabular}
    \vspace{-15pt}
\end{table*}

In this way, we can use BEV queries $Q$ to iterate over features from different places in the BEV space, time steps, multi-scale levels, and sampling heights. \textbf{The information from all over the places and all over the time can be directly retrieved without any loss in a unified manner}. This kind of design also makes long-range fusion possible since all features are directly accessed no matter how long before, which also enables adaptive temporal fusion.

The last major part of our method is the self-regression mechanism. Inspired by BEVFormer~\cite{li2022bevformer}, which concatenates the warped previous BEV features with the BEV queries before the self-attention module to realize the temporal fusion, we use a self-regression mechanism that concatenates the output of Transformer with the BEV queries as the new inputs and rerun the Transformer to get the final features. For the first running of the Transformer, we simply double and concatenate the BEV queries as the inputs. 

In BEVFormer, it is believed that the concatenation of warped BEV features and BEV queries brings temporal fusion, and it is the root cause of performance gain. In this work, we propose another explanation for this phenomenon, that is, the concatenation of BEV features and queries is to implicitly deepen and double the number of the Transformer's layers. Because the warped BEV features are already processed by the Transformer at previous time steps, the concatenation can be viewed as the grafting of two successive Transformers. In this way, a simple self-regression without warping can achieve a similar performance gain as BEVFormer. The detailed ablation study can be found in \cref{sec_ab}.
\vspace{-10pt}
\paragraph{Segmentation head}
We use a lightweight, fully convolutional model ERFNet~\cite{romera2017erfnet} as our segmentation head, which will upsample the output of the Transformer to the given BEV space resolution.
\section{Experiments}
\subsection{Dataset and Evaluation Settings}
\label{sec_eval}
\paragraph{Dataset} In this work, we use NuScenes~\cite{caesar2020nuscenes} as the evaluation dataset for the map segmentation task, which contains 1,000 driving scenes collected in Boston and Singapore. There are 28,130 and 6,019 keyframes for the training and validation set. Each keyframe contains six surrounding images.
\vspace{-10pt}
\paragraph{Evaluation settings} There are two widely used settings for the map segmentation task on NuScenes. The first one is the $100\text{m} \times 100\text{m}$ setting~\cite{philion2020lift,li2022bevformer,xie2022m} with two classes \texttt{road} and \texttt{lane}. The other one is the $60\text{m} \times 30\text{m}$ setting~\cite{li2021hdmapnet,peng2022bevsegformer,zhang2022beverse} with three classes \texttt{boundary}, \texttt{divider}, and \texttt{ped crossing}. In this work, we also propose a new $160\text{m} \times 100\text{m}$ setting for a more comprehensive evaluation, as shown in \cref{tb_setting}. The key motivations of the new setting are: 1) the evaluation range should be as large as the visible limit. 2) the evaluation criterion should be discriminative for both bad and good predictions. 3) the evaluation should avoid overfitting and show the ability of generalization\footnote{\label{ft}The detailed information, motivation, and derivation of the new setting can be found in the supplementary materials.}. In the new setting, we also use two difficulty levels ``easy'' and ``hard''. For the ``easy'' level, the evaluation is conducted with the front, rear, left, and right ranges of 50m, 30m, 30m, and 30m, respectively. The ``hard'' level is onducted with the left areas in the $160\text{m} \times 100\text{m}$ range.
For all settings, mean intersection-over-union (mIoU) is used as the evaluation metric.

\subsection{Implementation Details}
\label{sec_detail}
To evaluate the results of our method, we use ResNet50~\cite{he2016deep}, Swin-Tiny~\cite{liu2021swin}, and VoVNet~\cite{lee2019energy} as our backbones. The ResNet50 and Swin backbones are initialized from ImageNet~\cite{deng2009imagenet} pretraining, and VoVNet backbone is initialized from DD3D checkpoint~\cite{park2021pseudo}. The default number of layers of the Transformer is set to 12. The input image resolutions are set to $1600\times 900$ for ResNet50 and Swin. For VoVNet, we use $1408\times 512$ image size. We use AdamW~\cite{loshchilov2018decoupled} optimizer with a learning rate of 2e-4 and a weight decay of 1e-4. The learning rate is decreased by a factor of 10 for the backbone. The batch size is set to 1 per GPU, and models are trained with eight GPUs for 24 epochs. At the 20th epoch, the learning rate is decreased by a factor of 10. The number of multi-scale features is set to $L=4$, the default number of previous time steps is set to $P=6$, and the number of sampling heights is set to $Z=4$. The height range is $(-5\text{m},3\text{m}]$ with a stride of 2m.

For the $100\text{m} \times 100\text{m}$ setting, we use $50 \times 50$ BEV queries to represent the whole BEV space, then the results are upsampled by a factor of 4 to match the BEV resolution. For the $60\text{m} \times 30\text{m}$ setting, we use $100 \times 50$ BEV queries with a similar upsampling as the $100\text{m} \times 100\text{m}$ setting. 
For the $160\text{m} \times 100\text{m}$ setting, we use $80 \times 50$ BEV queries and then upsample 8x to match the ground truth resolution.
We use cross entropy loss to train on both settings. The loss weight for the background class is set to 0.4 by default for the class imbalance problem. Since the \texttt{road} class in the $100\text{m} \times 100\text{m}$ setting is polygon area without the class imbalance problem, the loss weight of the \texttt{road} background class is set to 1.0.

\subsection{Ablation Study}
\label{sec_ab}

\paragraph{Ability of long-range fusion} As discussed in the Introduction, the proposed unified multi-view fusion has the ability of long-range fusion since it can directly access both spatial and temporal information. In this part, We show the results of different fusion time steps to examine the ability of long-range fusion. 

\begin{figure}[h]
    \centering
    \includegraphics[width=0.9\linewidth]{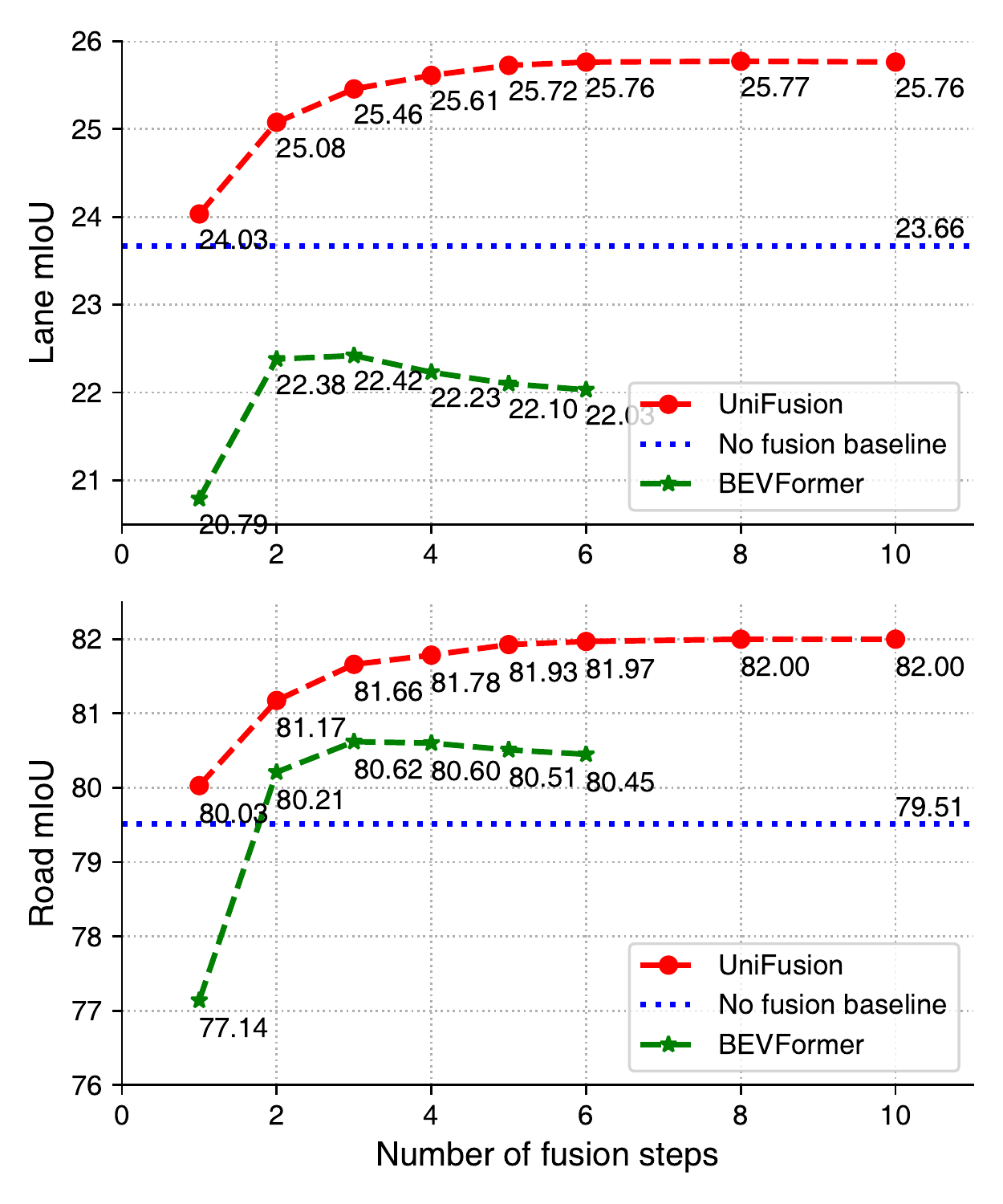}
    \vspace{-8pt}
    \caption{Ability of long-range temporal fusion.}
    \label{fig_long}
    \vspace{-10pt}
\end{figure}

From \cref{fig_long}, we can see that our method could consistently benefit from the long temporal fusion even up to 10 steps. And the fusion duration for the 10 steps is 2 seconds. However, the warp-based BEVFormer's performance would drop after 3 fusion steps. This is also in accord with the results in BEVFormer~\cite{li2022bevformer} that the performance of warp-based temporal fusion would decrease with longer fusion than 4 contiguous steps.
This shows the effectiveness of the proposed multi-view unified temporal fusion and the ability of long-range fusion.

Since the performance gradually converges after 6 fusion steps, we set the number of temporal fusion steps $P$ to 6 in this work.
\vspace{-10pt}
\paragraph{Disentangled training and inference fusion}
Although the proposed unified fusion has the ability of long-range fusion, this also brings another problem of computational complexity, especially during training. Longer fusion steps demand more memory and computational cost. We find a phenomenon that can alleviate this problem, \ie, the number of temporal fusion steps during training does not need to be the same as the one during inference. And a model trained with a short-range fusion setting still has the ability of long-range fusion during inference. We call this phenomenon disentangled training and inference fusion. The results are shown in \cref{tb_disentangled}.

\begin{table}[h]
    \centering
    \caption{Comparison of different numbers of temporal fusion steps. Note that the number of steps does not include current step. }
    \vspace{-8pt}
    \label{tb_disentangled}
    \setlength{\tabcolsep}{1mm}{
\begin{tabular}{cccc}
    \toprule
\begin{tabular}[c]{@{}c@{}}\#Fusion steps\\ (training) \end{tabular} & \begin{tabular}[c]{@{}c@{}}\#Fusion steps\\ (inference) \end{tabular} & Road mIoU & Lane mIoU \\ \midrule
0         & 0       &   79.04   &   22.64   \\
1         & 1       &   79.48   &   23.03   \\

1                                                                & 6                                                              & 81.12     & 24.24     \\
2                                                                & 6                                                              & 80.91     & 24.99     \\
3                                                                & 6                                                              & 81.02     & 24.48     \\
4                                                                & 6                                                              & 81.25     & 24.75    \\
\bottomrule
\end{tabular}
}
\vspace{-10pt}
\end{table}

From \cref{tb_disentangled}, we can see that no matter how many temporal fusion steps we use during training, the performance is very close when using 6 inference fusion steps. Moreover, even if we use only one previous step during training, the model still gains good performance with 6 temporal steps during inference. That is to say, the model still has the ability of long-range fusion when trained with a short-range fusion setting. By default, we use 2 temporal fusion steps during training.

\vspace{-10pt}
\paragraph{Effectiveness of self-regression mechanism} In \cref{sec_net}, we propose a self-regression mechanism to further boost the performance. In this part, we examine the effectiveness of the self-regression mechanism. As shown in \cref{tb_selfreg}, we can see that the model with self-regression always gains better performance. Interestingly, the performance of the 12-layer non-regression model is close to the one of the 6-layer self-regression model. This verifies the analysis in \cref{sec_net}. Moreover, we can see that the number of layers is also important for the final performance.
\begin{table}[h]
    \centering
    \caption{Comparison with different number of Transformer layers and self-regression.}
    \vspace{-8pt}
    \label{tb_selfreg}
    \setlength{\tabcolsep}{2mm}{
    \begin{tabular}{cc|cc}
    \toprule
    \#Layers & Self-Reg & Road mIoU & Lane mIoU \\ \midrule
    6        &          &    80.42       &     24.26      \\
    6        &    \checkmark      &    80.91       &    24.99       \\
    12       &          &           81.13  &  25.29         \\
    12       &    \checkmark      &    81.97       &    25.76      \\
    \bottomrule
    \end{tabular}
    }
    \vspace{-5pt}
\end{table}
\begin{table*}[t]
    \centering
    \caption{Experiments on NuScenes with the \textbf{$\mathbf{100\text{m} \times 100\text{m}}$} setting. * means the results are reported from BEVFormer~\cite{li2022bevformer}. $\dagger$ indicates that M2BEV uses a different setting, in which the BEV resolution is 2x larger. So the ``Lane mIoU'' is high.}
    \vspace{-8pt}
    \label{tb_100x100}
    \begin{tabular}{lccccccc}
    \toprule
    \multirow{2}{*}{Method}    & \multirow{2}{*}{Years}  & \multirow{2}{*}{Backbone} & \multirow{2}{*}{Parameters}         & \multirow{2}{*}{FPS} & \multicolumn{3}{c}{mIoU (Vanilla / City-based) }  \\   
     
        &   &  &  &               & Road mIoU & Lane mIoU & All\\ 
    \midrule
    LSS      & ECCV20 & EffNetb0 & -                     & -         & 72.9 / -\ \ \ \ \ \ \       & 20.0 / -\ \ \ \ \ \ \   &  46.5 / - \ \ \ \ \ \ \   \\

    VPN*      & IROS20 & Res101DCN & -                     & -         & 76.9 / -\ \ \ \ \ \ \       & 19.4 / -\ \ \ \ \ \ \      &  48.2 / - \ \ \ \ \ \ \ \\
    LSS*      & ECCV20 & Res101DCN & -                      & -         & 77.7 / -\ \ \ \ \ \ \       & 20.0 / -\ \ \ \ \ \ \        & 48.9 / - \ \ \ \ \ \ \  \\
    M2BEV     & -      & ResNeXt101 & 112.5                 & 1.4       & 77.2 / -\ \ \ \ \ \ \       & \ \ 40.5 / -$\dagger$\ \ \ \ \ \ \  & 58.9 / -$\dagger$ \ \ \ \ \ \hspace{1em}\\
    BEVFormer & ECCV22      & Res101DCN & 68.7            & 1.7       & 80.1 / -\ \ \ \ \ \ \       & 25.7 / -\ \ \ \ \ \ \     & 52.9 / - \ \ \ \ \ \ \    \\ 
    \midrule
    UniFusion & -      & ResNet50 & 42.4           & 2.6       & \textbf{82.0 / 42.6}      & \textbf{25.8 / 11.2}     & \textbf{53.9 / 26.9} \\
    UniFusion &       & VoVNet99 & 84.0            & 2.7       & \textbf{85.4 / 47.9}     & \textbf{31.0 / 11.6}    & \textbf{58.2 / 29.8} \\
    \bottomrule
    \end{tabular}
\end{table*}
\begin{table*}
    \centering
    \caption{Experiments on NuScenes with the \textbf{$\mathbf{60\text{m} \times 30\text{m}}$} setting. * means the results are reported from HDMapNet~\cite{li2021hdmapnet}. ** means the BEVFormer is reimplemented in this work.}
    \vspace{-8pt}
    \label{tb_60x30}
    \begin{tabular}{lcccccc}
    \toprule
    \multirow{2}{*}{Method} & \multirow{2}{*}{Years} & \multirow{2}{*}{Backbone} & \multicolumn{4}{c}{mIoU (Vanilla / City-based)}                 \\
                            &        &                 & Divider & Ped Crossing & Boundary & All  \\ \midrule
    VPN*                    & IROS20       & EffNetb0          & 36.5 / -\ \ \ \ \ \     & 15.8 / -\ \ \ \ \ \          & 35.6 / -\ \ \ \ \ \     & 29.3 / -\ \ \ \ \ \  \\
    LSS*                    & ECCV20       & EffNetb0          & 38.3 / -\ \ \ \ \ \     & 14.9 / -\ \ \ \ \ \          & 39.3 / -\ \ \ \ \ \      & 30.8 / -\ \ \ \ \ \  \\
    HDMapNet                & ICRA22       & EffNetb0          & 40.6 / -\ \ \ \ \ \     & 18.7 / -\ \ \ \ \ \          & 39.5 / -\ \ \ \ \ \      & 32.9 / -\ \ \ \ \ \  \\
    BEVSegFormer            & -            & ResNet101       & 51.1 / -\ \ \ \ \ \     & 32.6 / -\ \ \ \ \ \          & 50.0 / -\ \ \ \ \ \      & 44.6 / -\ \ \ \ \ \  \\
    BEVerse                 & -            & Swin-tiny          & 56.1 / -\ \ \ \ \ \     & \textbf{44.9} / -\ \ \ \ \ \         & 58.7 / -\ \ \ \ \ \      & 53.2 / -\ \ \ \ \ \  \\
    BEVFormer**                 & ECCV22            & ResNet50          & 53.0 / 20.4    & 36.6 / 8.9 \ \         & 54.1 / 24.3     & 47.9 / 17.9 \\
    \midrule
    UniFusion               & -            & Swin-tiny         & \textbf{58.6 / 32.4}    & 43.3 / 17.2         & \textbf{59.0 / 29.8}     & \textbf{53.6 / 26.5} \\
    UniFusion               & -            & VoVNet99          & \textbf{60.6 / 32.5}    & \textbf{49.0 / 11.5}         & \textbf{62.5 / 32.9}     & \textbf{57.4 / 25.6} \\ \bottomrule
    \end{tabular}
    \vspace{-10pt}
\end{table*}
\vspace{-10pt}
\paragraph{Unified cross attention brings adaptive temporal fusion} 
In \cref{eq_ca}, we show the core design of the unified multi-view spatial-temporal fusion is the unified cross attention module based on virtual views. The cross attention module can iterate over features from different time steps, which brings another important property, \ie, adaptive temporal fusion. To verify this, we directly average the $P$ temporal features before feeding them into the Transformer as the counterpart for comparison, which can be viewed as a fixed equal-weighted fusion. The results are shown in \cref{tb_adaptive}.

We can see that our method outperforms the equal-weighted temporal fusion counterpart in all settings. This shows that our method could adaptively fuse information from different time steps.

\vspace{-5pt}
\begin{table}[h]
    \centering
    \caption{Effectiveness of adaptive temporal fusion with different fusion steps. ``Avg.'' is the equal-weighted fusion.}
    \vspace{-5pt}
    \label{tb_adaptive}
    \setlength{\tabcolsep}{1mm}{
    \begin{tabular}{lcccccc}
    \toprule
    Fusion Steps    & 1     & 2     & 3     & 4     & 5     & 6     \\ \midrule
    UniFusion & 24.03 & 25.08 & 25.46 & 25.61 & 25.72 & 25.76 \\
    Avg.      & 23.26 & 24.47 & 24.82 & 24.95 & 25.03 & 25.08 \\
    \bottomrule
    \end{tabular}}
    \vspace{-10pt}
\end{table}

\subsection{Results}
To validate the performance of our method, we use VPN~\cite{pan2020cross}, Lift-Splat-Shoot~\cite{philion2020lift}, M2BEV~\cite{xie2022m}, and BEVFormer~\cite{li2022bevformer} for comparsion in the $100\text{m} \times 100\text{m}$ setting, as shown in \cref{tb_100x100}. The FPS of our method is measured on the RTX 3090 GPU.

We can see that the proposed method with a ResNet50 backbone even outperforms the BEVFormer model with a ResNet101DCN~\cite{dai2017deformable,wang2021fcos3d} backbone. In the \texttt{road} class, our method outperforms the previous SOTA BEVFormer by 1.9 points with the vanilla split. It is worth mentioning that BEVFormer uses much more BEV queries than ours ($200\times 200$ vs. $50 \times 50$), which could benefit the segmentation of thin lane lines. But our method still outperforms BEVFormer in the \texttt{lane} class with a smaller backbone and fewer BEV queries, which shows the effectiveness of the proposed UniFusion. Besides, our method also achieves the fastest speed compared with BEVFormer and M2BEV. Finally, our method with a larger VoVNet99 backbone outperforms BEVFormer by more than 5 points in all classes.

\begin{table*}
    \centering
    \caption{Comparison on NuScenes with the \textbf{$\mathbf{160\text{m} \times 100\text{m}}$} setting. We reimplement other methods with the same setting for comparison. All results are reported with the format of Vanilla split / City-based split.}
    \label{tb_160x100}
    \setlength{\tabcolsep}{0.7mm}{
    \begin{tabular}{lcc|cccc|cccc}
    \toprule
    \multirow{2}{*}{Method} & \multirow{2}{*}{Years} & \multirow{2}{*}{Backbone} & \multicolumn{4}{c|}{mIoU (Easy)} &  \multicolumn{4}{c}{mIoU (Hard)}             \\ 
                            &                        &                           & Divider & Crossing & Boundary & All & Divider & Crossing & Boundary & All \\ \midrule

    VPN                     & IROS20                 & ResNet50                  &  25.4 / 8.3\ \      & \ \ 6.7 / 0.5 \ \          & 25.3 / 14.6     &  19.1 / 7.8\ \   & 13.4 / 2.9     &  4.3 / 0.0          &  13.1 / 6.5     & 10.3 / 3.1 \\
    LSS                     & ECCV20                 & ResNet50                  &  11.3 / 6.4\ \      &  \ \ 0.3 / 0.2\ \         &  10.8 / 4.4\ \       &  7.5 / 3.7  & \ \ 6.0 / 1.2     &   0.4 / 0.2         &  6.2 / 1.1      &  4.2 / 0.8  \\
    BEVFormer               & ECCV22                      & ResNet50                  & 42.2 / 16.1    & 26.9 / 7.6\ \          & 42.1 / 18.6     & 37.1 / 14.1 & 27.3 / 7.8    & 17.5 / 2.3          & 26.3 / 10.0     & 23.7 / 6.7  \\
    UniFusion               & -                      & ResNet50                  & \textbf{46.3 / 18.5}    & \textbf{30.5 / 10.5}          & \textbf{45.8 / 21.0}     & \textbf{40.9 / 16.7} & \textbf{28.1 / 8.8}    & \textbf{17.6 / 2.7}          & \textbf{26.9 / 10.2}     & \textbf{24.2 / 7.2} \\

    \bottomrule
    \end{tabular}}
\end{table*}

\begin{figure*}
    \centering
    \includegraphics[width=0.85\linewidth]{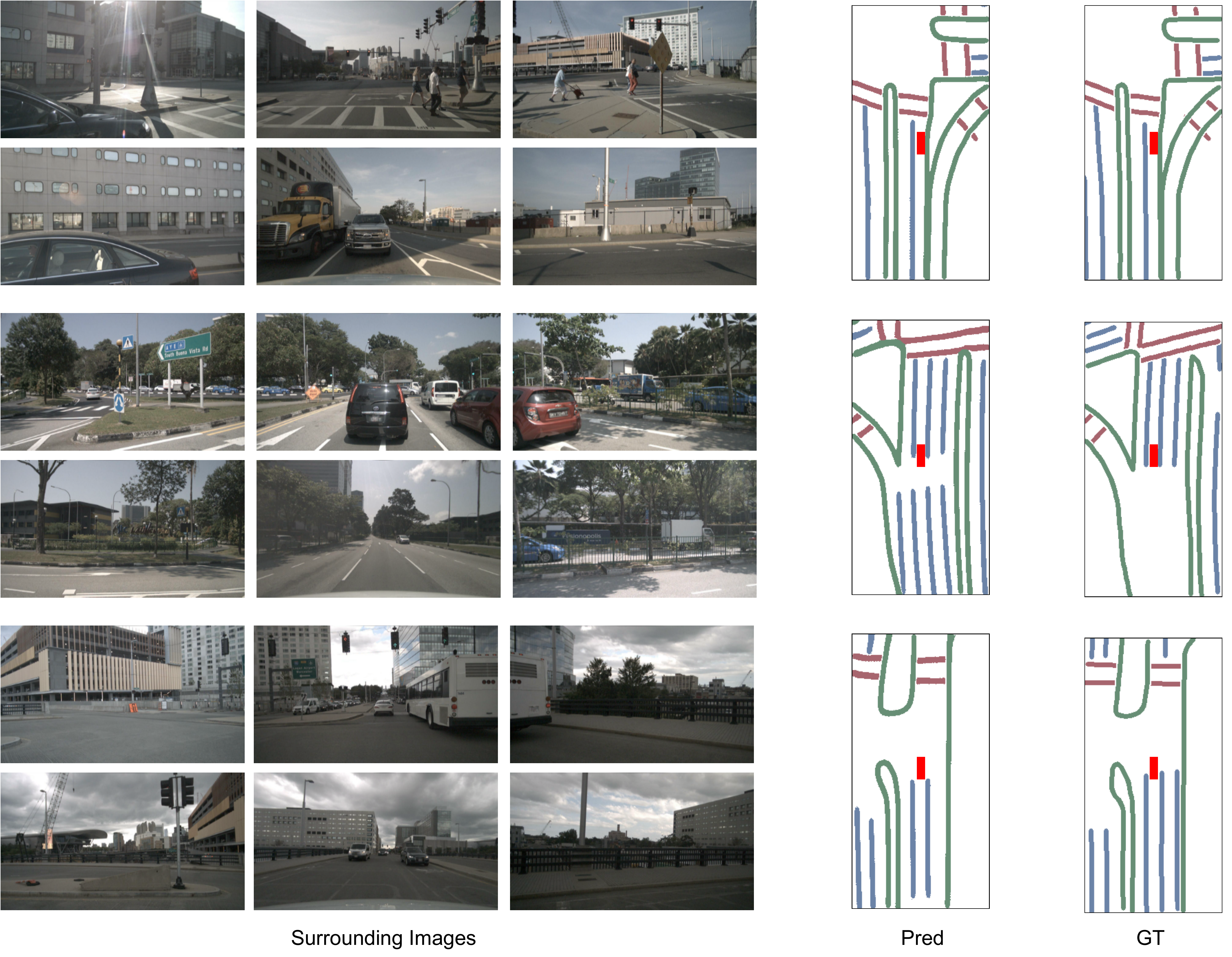}
    \caption{Visualization of our method on NuScenes \textit{val} set under complex road structures with the $60\text{m} \times 30\text{m}$ setting. From left to right, there are surrounding images, predictions, and ground truth. The red rectangle represents the ego car.}
    \label{fig_viz}
\end{figure*}

For the $60\text{m} \times 30\text{m}$ setting, we adopt VPN~\cite{pan2020cross}, Lift-Splat-Shoot~\cite{philion2020lift}, HDMapNet~\cite{li2021hdmapnet}, BEVSegFormer~\cite{peng2022bevsegformer}, and BEVerse~\cite{zhang2022beverse} for comparsion. The comparison results are shown in \cref{tb_60x30}. From \cref{tb_60x30}, we can see that our method still obtains the best results in all settings.

In order to better evaluate different models and provide a scenario that is closer to real-world autonomous driving, we also introduce a new $160\text{m} \times 100\text{m}$ setting. We use VPN~\cite{pan2020cross}, LSS~\cite{philion2020lift}, BEVFormer~\cite{li2022bevformer}, and our method with the same training setting for comparison, as shown in \cref{tb_160x100}. 

From \cref{tb_160x100} we can see that visible range is crucial for the map segmentation task. And the relatively low performance suggests that large-range real-world map segmentation is still an open problem. Finally, we can see our method still obtains the best performance.

It should be noted that the vanilla NuScenes \textit{train}/\textit{val} sets contain many similar samples, and it is likely to be influenced by overfitting. In this way, we introduce the new city-based split for NuScenes, the results can be seen in \cref{tb_100x100,tb_60x30,tb_160x100}. We can see that with the city-based split, all methods' performance drops significantly, and the poor improvement of VoVNet in \cref{tb_60x30} with the city-based split also indicates the problem of overfitting. This could be an important direction for future works.

At last, we show the visualization results of our method, as shown in \cref{fig_viz}. We can see that our method gains good results under complex road structures. Our method could even segment the parts that are missing in the ground truth, as shown in the second row. Moreover, for the irregular road boundary, our method still gains good results.

\section{Conclusion}
In this work, we propose a unified spatial-temporal fusion method for BEV representation, termed UniFusion. Different from previous methods that use warpping, we propose a new concept, \ie, virtual views that merge both spatial and temporal fusion in a unified formulation. With this design, we can realize long-range and adaptive temporal fusion with no information loss. The experiments and visualizations validate the effectiveness of our method.

{\small
\bibliographystyle{ieee_fullname}
\bibliography{egbib}
}
\newpage
\onecolumn
\setcounter{section}{0}
\setcounter{figure}{0}
\setcounter{table}{0}
\setcounter{equation}{0}
\renewcommand\thetable{\Alph{table}}
\renewcommand\thefigure{\Alph{figure}}

\large{\textbf{\centerline{Supplementary Materials}}}

\section{Overview}
In this part, we provide more detailed illustration, explanation, and visualization for the following aspects: 1) Comparison under different computational costs. 2) The motivation of the new $160\text{m} \times 100\text{m}$ setting; 3) The long-range fusion ability of warp-based methods. 3) Visual comparison of different methods.

\section{Comparison under different computational costs}
\begin{figure}[h]
    \centering
    \includegraphics[width=0.6\textwidth]{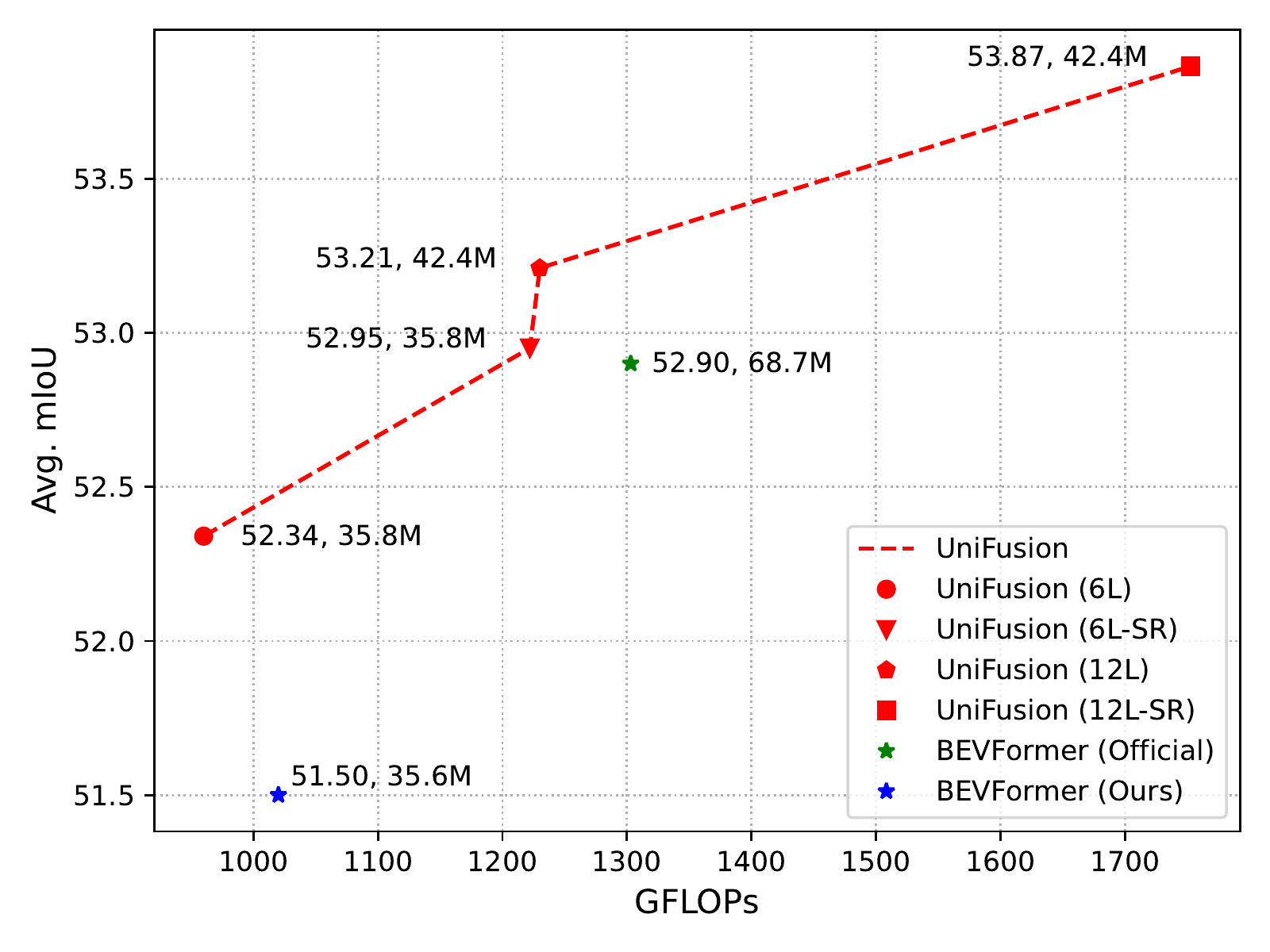}
    \caption{FLOPs vs. performance. The variants of UniFusion are derived by adjusting the number of layers in the fusion Transformer and whether the self-regression is utilized. “6L” means the Transformer is 6-layer. “SR” means self-regression.}
    \label{fig_flops}
\end{figure}

Although our method could support long-range temporal fusion and gains better performance, it would has a higher computational cost compared with the short-range temporal fusion methods. For fair comparison, we scale our method's computational costs to compare with BEVFormer, as shown in \cref{fig_flops}. It should be note that we only scale the Transformer module which is used for fusion. All other settings like backbone, input resolution, training settings and task-specific head remain unchanged.

From \cref{fig_flops} we can see that the proposed method could outperforms BEVFormer with lower computational costs and parameters. This shows that the proposed method could not only support long-range temporal fusion, but also has a high efficiency.

\section{Motivation of the $160\text{m} \times 100\text{m}$ setting}
Generally speaking, we propose a new $160\text{m} \times 100\text{m}$ setting that has different BEV range, line width of map element, and split compared with the existing $60\text{m} \times 30\text{m}$ and $100\text{m} \times 100\text{m}$ settings. The key motivations of this setting are: 1) the evaluation range should be as large as the visible limit. 2) the evaluation criterion should be discriminative for both bad and good predictions. 3) the evaluation should avoid overfitting and show the ability of generalization.
\subsection{BEV Range}
To determine the BEV range, we consider the visible limit of cameras. In this work, we define the visible range as the farthest point where a lane is represented by less than two pixels in the feature map (since we need to distinguish the left and right lanes of the lane, two pixels is the minimum requirement). Suppose $f$ is the focal length of the camera, $n_{pixel}$ is the minimal number of pixels to represent a lane, and $W_{lane}$ is the width of the lane. The visible limit $d$ can be written as:
\begin{equation}
    d = \dfrac{f}{n_{pixel}} W_{lane}
\end{equation}
\begin{figure*}[h]
    \centering
    \includegraphics[width=0.7\linewidth]{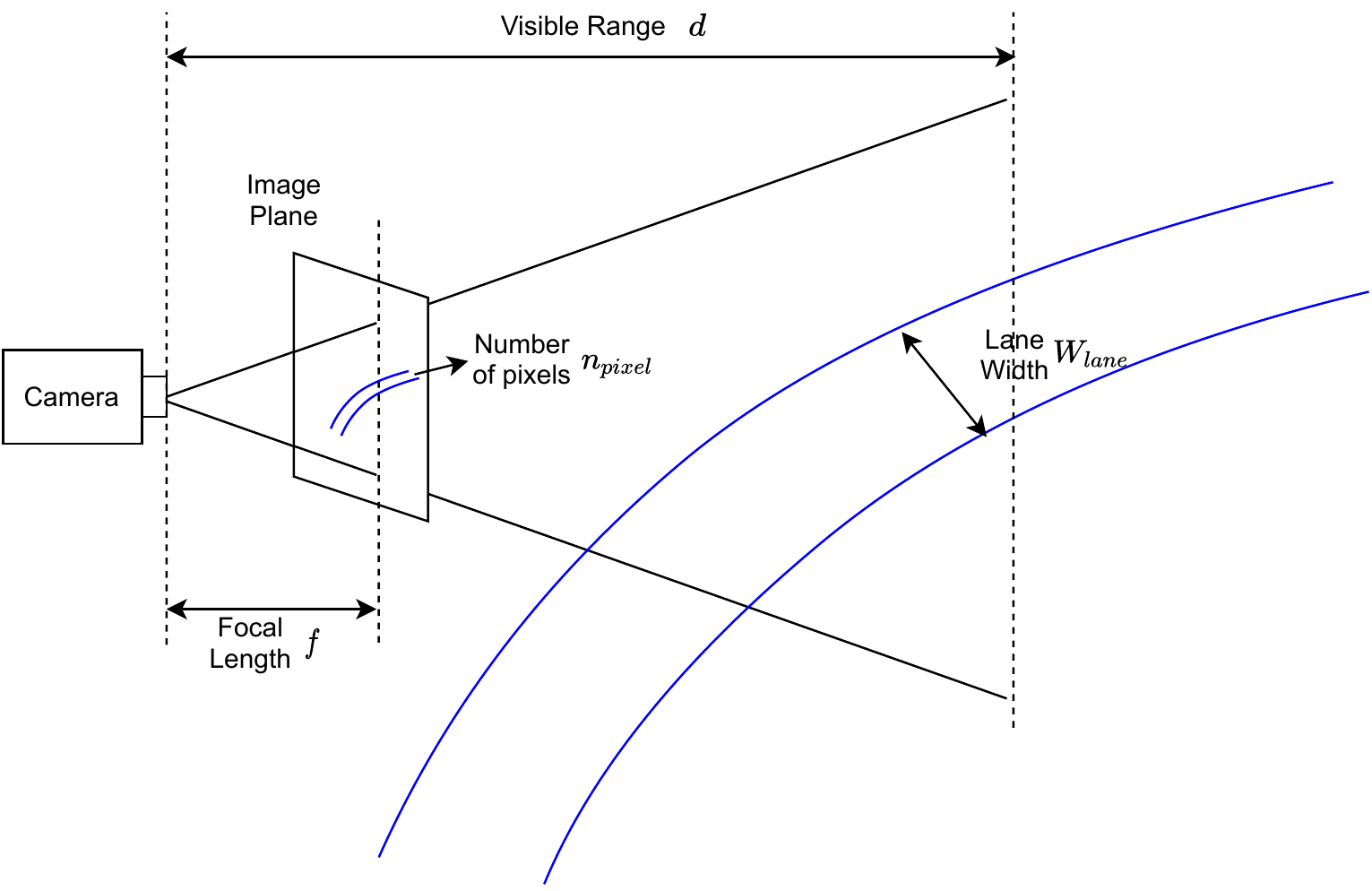}
    \caption{Derivation of BEV range.}
    \label{fig_bev_range}
\end{figure*}
An example of the derivation is shown in \cref{fig_bev_range}. Typically, the focal length on NuScenes can be derived from the FOV and image resolution. Suppose image resolution is $r$, FOV is $\theta$, and we have:
\begin{equation}
    f = \frac{r/2}{\text{tan}(\theta/2)}
\end{equation}
\begin{table}[h]
    \centering
    \caption{The values on the NuScenes dataset. For the FOV and focal length, we list the values of front and rear cameras separately. Lane width is about 3.0m-4.0m according to the regulations of different places, and we use the minimum value of 3.0m. Since the common network output stride is larger than 32, one pixel in the feature map corresponds to at least 32 pixels in the original image.}
    \label{tb_value}
    \begin{tabular}{ccccc}
    \toprule
    Image Resolution $r$ & FOV $\theta$   & Focal Lenght $f$ & Lane Width $W_{lane}$ & Number of pixels $n_{pixel}$ \\ \midrule
    1600             & 70 / 110 & 1142.5 / 560.2 & 3.0m & 32 \\
    \bottomrule
    \end{tabular}
\end{table}
The detailed numbers are shown in \cref{tb_value}. 

Finally, we get the BEV range $d$:
\begin{equation}
    \begin{aligned}
    d_{front} =\  \dfrac{1142.5}{32} \cdot 3 &\approx 107.1 \\
    d_{rear} =\  \dfrac{560.2}{32} \cdot 3 &\approx 52.5
    \end{aligned}
\end{equation}
However, the rear BEV range of 52.5m is slightly short in real scenarios. We slightly extend the rear BEV range to 60m. For the left and right range, we follow the existing setting with a distance of 50m. This composes the $160\text{m} \times 100\text{m}$ setting.


\subsection{Evaluation criterion}
The first difference in the evaluation criterion is that all the map elements are defined as the ``Line''. This is because the polygon area is not suitable for representing road structures and the mIoU metric with polygon is abnormally high. For example, the ``Road mIoU'' is about 80 while the ``Lane mIoU'' is only about 20.

The second part of our evaluation is the line width. In this work, we use 3-pixel-wide lines. This is to avoid the problem of the 1-pixel evaluation. For example, if the predicted lane is only shifted by 1 pixel from the ground truth, then the mIoU is 0. There is no discrimination for ``wrong but close'' and `totally wrong`'' cases under this setting. This property also causes another problem, that is, if we simply upsample the ground truth and make the prediction also works in high resolution, the performance would increase significantly, which would cause an unfair comparison between different methods. To avoid these problems, we set the line width to 3 pixels. For the predictions that are close to ground truth but not exactly correct, our evaluation could also give responses to these results and are more discriminative. For the upsample problem, since we make the original 1-pixel ``lane mIoU'' a 3-pixel ``area mIoU'', the upsampled results are less affected.

\subsection{City-based split}

In our setting, we also propose the city-based split for NuScenes. This is because the vanilla training and validation splits in NuScenes contain many similar scenes, which potentially suffer from the overfitting problem. In this way, we propose a split that is based on the cities and locations on NuScenes. NuScenes is collected in four places, which are ``singapore-onenorth'', ``singapore-queenstown'', ``singapore-hollandvillage'', and ``boston-seaport''. We use the samples collected in ``singapore-queenstown'' and ``singapore-hollandvillage'' as the training split, and ``singapore-onenorth'' and ``boston-seaport'' as the validation split. The numbers of training and validation samples are 26,093 and 8,056, respectively. For comparison, the numbers of training and validation samples in the vanilla split are 28,130 and 6,019, respectively. The detailed split list can be found in \url{https://github.com/cfzd/UniFusion}.

\section{Visualization and Comparison}
\begin{figure*}[h]
    \centering
    \includegraphics[width=1.\linewidth]{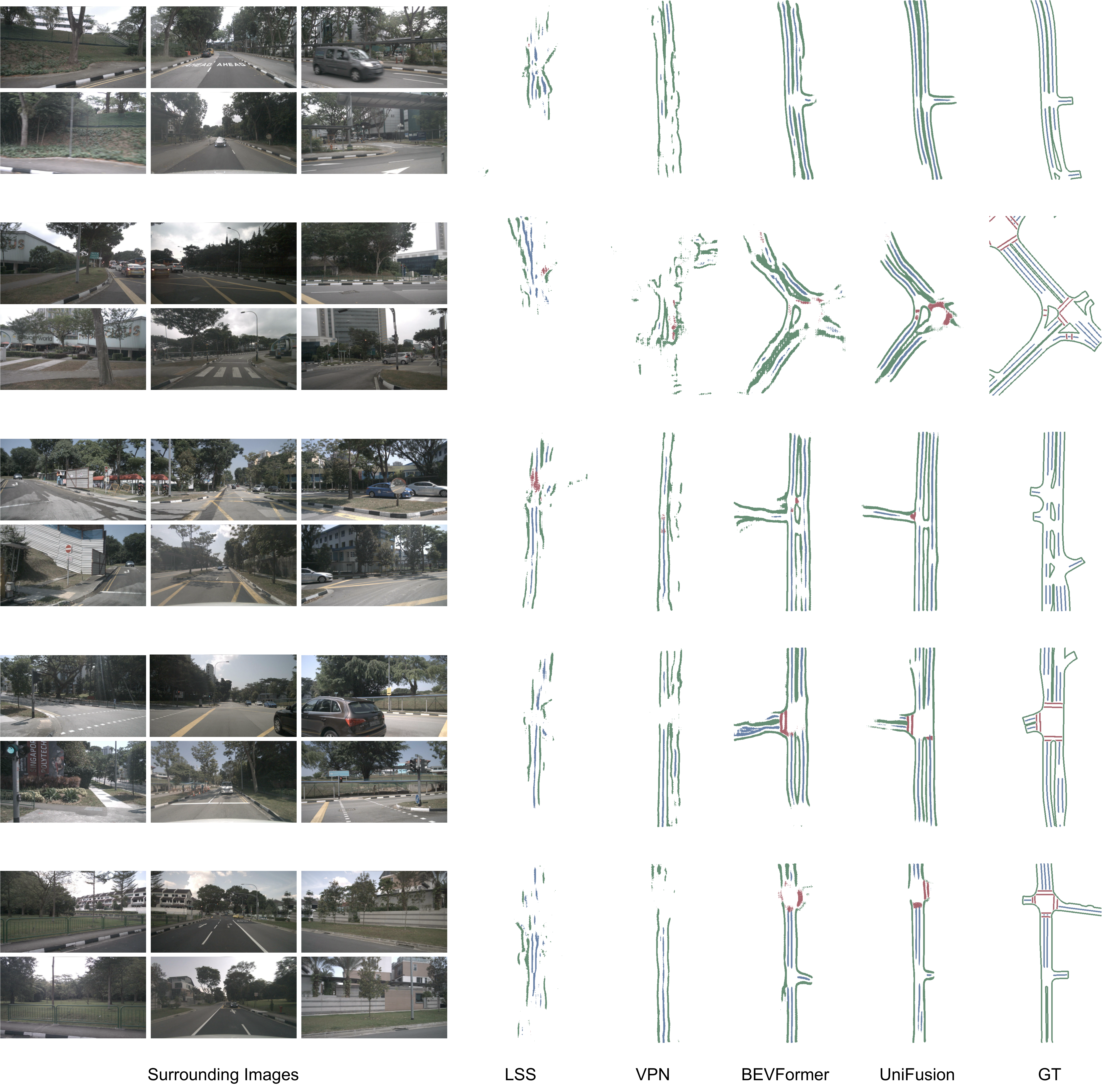}
    \caption{The visual comparison on the city-based val split of NuScenes with the $160\text{m} \times 100\text{m}$ setting. Best viewed when zoomed in.}
    \label{tb_supp_viz}
\end{figure*}

In this part, we show the visualization results on NuScenes with the $160\text{m} \times 100\text{m}$ setting. Moreover, we also show the results of other method for comparison in \cref{tb_supp_viz}. From \cref{tb_supp_viz}, we can see that our method gains the best results. The prediction of lines in our method is smooth and clear.

\end{document}